\documentclass{article}

\usepackage{ijcai17}

\usepackage{times}
\usepackage{algorithm}
\usepackage{algorithmic}
\usepackage{amsmath,amsfonts,amssymb}
\usepackage{hyperref}
\usepackage{graphicx}
\usepackage{sidecap}
\usepackage{caption}
\usepackage{subcaption}
\usepackage{verbatim}
\usepackage[para]{footmisc}
\usepackage{enumitem}
\usepackage{amsmath}	
\usepackage{multirow}
\usepackage{makecell}
\sidecaptionvpos{figure}{t}

\newcommand\numberthis{\addtocounter{equation}{1}\tag{\theequation}}

\newcommand{\mtx}[1]{\ensuremath{\mathbf{#1}}}

\newcommand{\usparse}{\mtx{r_{i,.}}}
\newcommand{\vsparse}{\mtx{r_{.,j}}}
\newcommand{\uestim}{\mtx{\hat{r}_{i,.}}}
\newcommand{\vestim}{\mtx{\hat{r}_{.,j}}}

\DeclareMathOperator*{\argmin}{arg\,min} % Jan Hlavacek

\begin{document}

%\title{Hybrid Matrix Completion with Autoencoders}
%\title{Efficient Matrix Completion with Autoencoders}
\title{Hybrid Recommender System based on Autoencoders}

\author{Florian Strub \\ \small florian.strub@inria.fr\normalsize \\ \small Univ. Lille, CNRS, Centrale Lille, Inria \And J\'er\'emie Mary \\ \small jeremie.mary@univ-lille3.fr\normalsize \\ \small Univ. Lille, CNRS, Centrale Lille, Inria \And Romaric Gaudel \\ \small romaric.gaudel@univ-lille3.fr\normalsize \\ \small Univ. Lille, CNRS, Centrale Lille, Inria }
%\institute{CRIStAL - SequeL Team, Univ. Lille,
% 40 av. du Halley
%59650 Villeneuve d'Ascq FRANCE}

\maketitle

\begin{abstract}
Proficient Recommender Systems heavily rely on Matrix Factorization (MF) techniques. MF aims at reconstructing a matrix of ratings from an incomplete and noisy initial matrix; this prediction is then used to build the actual recommendation. Simultaneously, Neural Networks (NN) met tremendous successes in the last decades but few attempts have been made to perform recommendation with autoencoders. In this paper, we gather the best practice from the literature to achieve this goal. We first highlight the link between these autoencoder based approaches and MF. Then, we refine the training approach of autoencoders to handle incomplete data. Second, we design an end-to-end system which handles external information. Finally, we empirically evaluate these approaches on the MovieLens and Douban dataset. 
\end{abstract}

\section{Introduction}

Recommendation systems advise users on which items (movies, music, books etc.) they are more likely to be interested in. A good recommendation system may dramatically increase the number of sales of a firm or retain customers. For instance, 80\% of movies watched on Netflix come from the recommender system of the company \cite{Netflix2015}. Collaborative Filtering (CF) aims at recommending an item to a user by predicting how a user would rate this item. To do so, the feedback of \emph{one} user on \emph{some} items is combined with the feedback of \emph{all} other users on \emph{all} items to predict a new rating. For instance, if someone rated a few books, CF objective is to estimate the ratings he would have given to thousands of other books by using the ratings of all the other readers.
The most successful approach in CF is to factorize an incomplete matrix of ratings \cite{Koren2009,Zhou2008}. This approach simultaneously learns a representation of users and items that encapsulate the taste, genre, writing styles, etc.
%Common latent factor techniques compute a low-rank approximation of the rating matrix by different means \cite{Koren2009,Zhou2008}. 
%However, these methods are linear and may not catch subtle latent factors. Newer algorithms were explored to face those constraints such as \cite{FIND_ONE}.
%More recent works combine several low-rank matrices into a single one to improve the quality of the final matrix \cite{Lee2013}.
A well-known limit of this approach is the \emph{cold start} setting: how to recommend an item to a user when few rating exists for either the user or the item?

While Deep Learning methods start to be used for several scenarios in recommendation system that goes from dialogue systems~\cite{wen2016network} to temporal user-item interactions~\cite{dai2016recurrent} through heterogeneous classification applied to the recommendation setting~\cite{cheng2016wide}, few attempts have been done to use Neural Networks (NN) in CF. Deep Learning key successes have mainly been on fully observable data~\cite{Lecun2015} while CF relies on input with missing values. This constraint has received less attention and remains a challenging problem for NN.
Handling missing values for CF has first been tackled by Salakhutdinov et al. \cite{Salakhutdinov2007} for the Netflix challenge by using Restricted Boltzmann Machines. More recent works train autoencoders to perform CF tasks on explicit data \cite{Sedhain2015,Strub2015} and implicit data \cite{Zheng2016}. Other approaches rely on recurrent networks~\cite{Wang2016collaborative}. Although those methods report excellent results, they ignore the cold start scenario and provide little insights.

The key contribution of this paper is to collect the best practices of autoencoder approaches in order to standardize the use of autoencoders for CF tasks.
We first highlight autoencoders perform a factorization of the matrix of ratings.
Then we contribute to an end-to-end autoencoder based approach in two points: (i) we introduce a proper training loss/process of autoencoders on incomplete data, (ii) we integrate the side information to autoencoders to alleviate the cold start problem. 
Compared to previous attempts in that direction \cite{Salakhutdinov2007,Sedhain2015,Wu2016}, our framework integrates both ratings and side information into a unique network. This joint model leads to a scalable and robust approach which beats state-of-the-art results in CF.
Reusable source code is provided in Lua/Torch to reproduce the results.

The paper is organized as follows. First, Sec.~\ref{sec:preliminaries} fixes the setting and highlights the link between autoencoders and matrix factorization in the context of CF. Then, Sec.~\ref{sec:model} describes our model. Finally, Sec.~\ref{sec:exp} details several experimental results from our approach.

\section{State of the art}
\label{sec:preliminaries}

\subsection{Matrix Factorization based Collaborative Filtering}
One of the most successful approach of CF consists of completing the matrix of ratings through Matrix Factorization (MF) \cite{Koren2009}.
Given $N$ users and $M$ items, we denote $r_{ij}$ the rating given by the $i^{th}$  user for the $j^{th}$ item. 
It entails an incomplete matrix of ratings $\mtx{R} \in \mathbb{R}^{N\times M}$. 
MF aims at finding a rank $k$ matrix $\mtx{\widehat{R}} \in \mathbb{R}^{N\times M}$ which matches known values of $\mtx{R}$ and predicts unknown ones. Typically, $\mtx{\widehat{R}} = \mtx{U}\mtx{V}^{T}$ with $\mtx{U} \in \mathbb{R}^{N\times k}$, $\mtx{V} \in \mathbb{R}^{M\times k}$ and  ($\mtx{U}$~,$\mtx{V}$) the solution of
  $$\argmin_{\mtx{U},\mtx{V}}
  \sum_{(i,j) \in \mathcal{K}(\mtx{R}) }  (r_{ij} - \mtx{u}_i^T\mtx{v}_j)^2 + \lambda( \|\mtx{u}_i \|_{F}^2 + \|\mtx{v}_j \|_{F}^2),$$
where 
 $\mathcal{K}(\mtx{R})$ is the set of indices of known ratings of $\mtx{R}$,
 ($\mtx{u}_i$, $\mtx{v}_j$) are column-vectors of the low rank rows of ($\mtx{U}$, $\mtx{V}$) and 
 $\|.\|_{F}$ is the Frobenius norm.
In the following, $\usparse$ and $\vsparse$ will respectively be the $i$-th row  and $j$-th column of $\mtx{R}$.

\subsection{Autoencoder based Collaborative Filtering}

Recently, autoencoders have been used to handle CF problems \cite{Sedhain2015,Strub2015}.
Autoencoders are NN popularized by Kramer \cite{Kramer1991}.% which handle an unsupervised task: reconstruct the input on the output of the network.
They are unsupervised networks where the output of the network aims at reconstructing the input.
The network is trained by back-propagating the squared error loss on the output.
More specifically, when the network limits itself to one hidden layer, its output is
$$ nn(\mtx{x}) \stackrel{\text{def}}{=} \sigma(\mtx{W_2 \sigma(\mtx{W_1 x + b_1}) + b_2}),$$
with $\mtx{x} \in \mathbb{R}^{N}$ the input, $\mtx{W_1} \in
\mathbb{R}^{k\times N}$ and $\mtx{W_2} \in \mathbb{R}^{N\times k}$ the weight
matrices, $\mtx{b_1} \in \mathbb{R}^{k}$ and $\mtx{b_2} \in \mathbb{R}^{N}$ the
bias vectors, and $\sigma(.)$ a non-linear transfer function.
%The size $k \ll N$ of the hidden layer is called the bottleneck.
%The network is finally trained by back-propagating the squared error loss on the reconstruction. 
%
%Autoencoders were introduced to identify correlations in problem variables as an aid to dimensionality reduction \cite{Kramer1991}.

In the context of CF, the autoencoder is fed with incomplete rows $\usparse$ (resp. columns $\vsparse$) of $\mtx{R}$ \cite{Sedhain2015,Strub2015}. It then outputs a vector $\uestim$ (resp. $\vestim$) which predict the missing entries.
Note that these approaches perform a non-linear low-rank approximation of $\mtx{R}$.
Using MF notations and assuming that the network works on rows $\usparse$ of $\mtx{R}$, we recover a predicted vector $\uestim$ of the form $\uestim=\sigma\left(\mtx{V}\mtx{u_i}\right)$:
\begin{equation*}
\uestim = nn(\usparse)
= \sigma \left(\; \underbrace{\left[\mtx{W_2} \; {\mtx{I}_N}\right]}_{\mtx{V}}\;
\underbrace{\left[\begin{array}{c}\sigma(\mtx{W_1\usparse + b_1})\\\mtx{b_2}\end{array}
\right]}_{\mtx{u_i}} \;\right).
\end{equation*}
The activation of the hidden units of the autoencoder $\mtx{u_i}$ iteratively builds the low rank matrix $\mtx{U}$. Besides, the final matrix of weights corresponds to the low rank matrix $\mtx{V}$. In the end, the output of the autoencoder performs a non linear matrix factorization $\hat{\mtx{R}} = \sigma\left(\mtx{U}\mtx{V}^T\right)$. Identically, it is possible to iterate over the columns $\vsparse$ of $\mtx{R}$ to compute $\vestim$ and get the final matrix $\hat{\mtx{R}}$.

%Note that CFN also breaks the symmetry between $\mtx{U}$ and $\mtx{V}$. For example, while Matrix Factorization approaches learn both $\mtx{U}$ and $\mtx{V}$, U-CFN learns $\mtx{V}$ and only indirectly learns $\mtx{U}$: U-CFN targets the function to build $\mtx{u_i}$ whatever the row $\usparse$. A nice benefit is that the learned autoencoder is able to fill in every vector $\usparse$, even if that vector was not in the training data. Both non-linear decompositions on rows and columns are done independently, which means that the matrix $\mtx{V}$ learned by U-CFN from rows can differ from the concatenation of vectors $\mtx{v_j}$ predicted by I-CFN from columns. 

\begin{figure*}[t]
\centering
\includegraphics[width=0.8\linewidth]{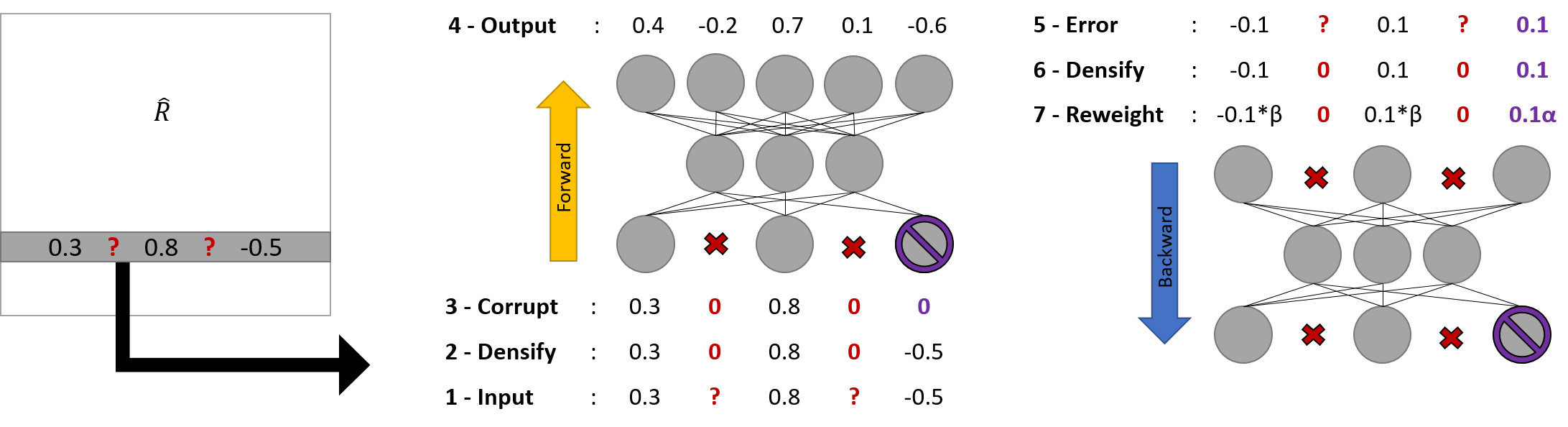}
\caption{Training steps for autoencoders with incomplete data. The input is drawn from the matrix of ratings, unknown values are turned to zero, some ratings are masked (corruption) and a dense estimate is obtained. Before backpropagation, unknown ratings are turned to zero error, prediction errors are reweighted by $\alpha$ and reconstruction errors are reweighted by $\beta$.}
\label{img:training}
\vskip -0.5em
\end{figure*}

%\subsection{Related Work}
\subsection{Challenges}
While the link between MF and autoencoders is straightforward, there is no guarantee that autoencoders can successfully perform matrix completion from incomplete data. Indeed, most of the prominent results with autoencoders such as word2vec \cite{Mikolov2013} only deal with complete vectors. The task with missing data is all the more difficult as the missing data have an impact on both input and target vectors.
%
%A few approaches exist in the literature to deal with missing values in neural networks. First, the problem is sometimes simply avoided by pre-computing an estimate of the missing values \cite{Tresp1994}. Yet, this solution is not satisfactory in our case as we want the autoencoder to handle this prediction issue itself.
Miranda et al.~\cite{Miranda2012} study the impact of missing values on autoencoders with industrial data. Yet,
they % Vladimiro Miranda est un homme : https://sigarra.up.pt/feup/en/func_geral.formview?p_codigo=208389
only have 5\% of missing values in her dataset, whereas CF tasks usually have more than 95\% of missing values.

Autorec \cite{Sedhain2015} handles the missing values by associating one autoencoder per sample, whose input size matches the number of known values. The corresponding weight matrices are shared among the autoencoders. Even if this approach is intellectually appealing, it faces technical limitations. First, sharing weights among networks prevent efficient computations (especially on GPU). Secondly, it prevents gradient optimization methods such as momentum, weight decay, etc. from being applied to missing data.
%
%Before that, Restrictive Boltzmann Machines (RBM) \cite{Salakhutdinov2007} has been used for CF during the Netflix Prize \cite{Netflix2015}. Although autoencoders and RBM are historically linked, RBM only handles binary inputs. Therefore, it strictly reduces the use of RBM on dataset with real numbers. Besides, the  weight architecture of RBMs discards the direct link between matrix factorization and RBM.
%
In the rest of the paper, we introduce a new method based on a single autoencoder to fully regain the benefits of NN techniques. Although this architecture is mathematically equivalent to a mixture of autoencoders \cite{Sedhain2015}, it turns out to be a more flexible framework.

CF systems also face the \emph{cold start} problem. The main solution is to supplant the lack of ratings by integrating side information. Some approaches \cite{Burke:2002} mix CF with a second system based only on side information. Recent work tends to incorporate the side information into the matrix completion by modifying the training error \cite{Adams2010,Chen2012,Rendle2010,Porteous2010}. 
%However, the training error are often hand-crafted and can be hard to fine-tune.
%
In this line of research, some papers incorporate NN embedding on side information. For instance, \cite{Wang2014} respectively auto-encode bag-of-words from movie plots, \cite{Li2015} auto-encode heterogeneous side information from users and items. Finally, \cite{Wang2014b} uses convolutional networks on music samples. From the best of our knowledge, training a NN from end to end on both ratings and heterogeneous side information has never been done. In Sec.~\ref{sec:sideinfo}, we build such NN by integrating side information into our CF system.

\section{End-to-End Collaborative Filtering with Autoencoders }
\label{sec:model}

Our approach builds upon an autoencoder to predict full vectors $\uestim$/$\vestim$ from incomplete vectors $\usparse$/$\vsparse$. As in \cite{Salakhutdinov2008,Sedhain2015,Strub2015}, we define two types of autoencoders: U-CFN is defined as $\uestim = nn(\usparse)$ and predicts the missing ratings given by the users; I-CFN is defined as $\vestim = nn(\vsparse)$ and predicts the missing ratings given to the items. First, we design a training process to predict missing ratings from incomplete vectors. Secondly, we extend CF techniques using side information to autoencoders to improve predictions for users/items with few ratings.

\subsection{Handling Incomplete Data}

MC tasks introduce two major difficulties for autoencoders. The training process must handle incomplete input/target vectors. The other challenge is to predict missing ratings as opposed to the initial goal of autoencoders to rebuild the initial input. We handle those obstacles by two means. 

First, we inhibit input and output nodes corresponding to missing data. For input nodes, the inhibition is obtained by setting their value to zero. To inhibit the back-propagated unknown errors, we use an empirical loss that disregards the loss of unknown values. No error is back-propagated for missing values, while the error is back-propagated for actual zero values.

Second, we shake the training data and design a specific loss function to enforce reconstruction of missing values. Indeed, basic autoencoders loss function only aims at reconstructing the initial input. Such approach misses the point that CF goal is to predict missing ratings. Worse, there is little interest to reconstruct already known ratings. To increase the generalization properties of autoencoders, we apply the Denoising AutoEncoder (DAE) approach \cite{Vincent2008} to the CF task we handle.
The DAE approach corrupts the input vectors and lets the network denoise the outputs. The corresponding loss function reflects that objective and is based on two main hyperparameters: $\alpha$ and $\beta$. They balance whether the network would focus on denoising the input ($\alpha$) or reconstructing the input ($\beta$). In our context, the data are corrupted by masking a small fraction of the known rating.
This corruption simulates missing values in the training process to train the autoencoders to predict them.
We also set $\alpha > \beta$ to emphasize the prediction of missing ratings over the reconstruction of known ratings. 
In overall, the training loss with regularization is:
\begin{multline}\label{eq:loss}
{\cal L}_{\alpha, \beta}(\mtx{x},\mtx{\tilde{x}}) = 
\alpha\left(\sum_{j\in \mathcal{K}(\mtx{x}) \cap \mathcal{C}(\tilde{\mtx{x}})}(nn(\mtx{\tilde{x}})_j - x_j)^2\right)
 + \\
\beta \left(\sum_{j\in \mathcal{K}(\mtx{x}) \setminus \mathcal{C}(\tilde{\mtx{x}})}(nn(\mtx{\tilde{x}})_j - x_j)^2\right) + \lambda\|\mtx{W}\|_{F}^2,
\end{multline}
where \mtx{\tilde{x}} is the corrupted input, $\mathcal{C}$ contains the indices of corrupted elements in $\mtx{\tilde{x}}$, $\mathcal{K}(\mtx{x})$ contains the indices of known values of $\mtx{x}$, $\mtx{W}$ is the flatten vector of weights of the network and $\lambda$ is the regularization parameter. Note that the regularization is applied to the full matrix of weights as opposed to \cite{Sedhain2015}. The overall training process is described in Fig.~\ref{img:training}.

\subsection{Integrating Side Information}
\label{sec:sideinfo}

CF only relies on the feedback of the users regarding a set of items. However, adding more information may help to increase the prediction accuracy and robustness. Furthermore, CF suffers from the cold start problem: when little information is available on an item, it will greatly lower the prediction accuracy. We now integrate side information to CFN to alleviate the cold start scenario. %This is process is known as hybridization in the recommendation community. 

The simplest approach to integrate side information to MF techniques is to append a user/item bias to the rating prediction \cite{Koren2009}: 
$\hat{r}_{ij} = \mtx{u}_i^T\mtx{v}_j + b_{u,i} + b_{v,j} + b',$
where $b_{u,i}$, $b_{v,j}$, $b'$ are respectively the user, item, and global bias. These biases are computed with hand-crafted engineering or CF technique \cite{Chen2012,Porteous2010,Rendle2010}. For instance, side information can directly be concatenated to the feature vector $\mtx{u}_i$/$\mtx{v}_j$ of rank $k$. Therefore, the estimated rating is computed by:
\begin{align}\label{eq:side_info1}
\hat{r}_{ij} &= \{\mtx{u}_i, \mtx{x}_i\} \otimes \{\mtx{v}_j, \mtx{y}_j\} \\
 \stackrel{\text{def}}{=} &
\mtx{u}_{[1:k], i}^T\mtx{v}_{[1:k], j} + 
\underbrace{\mtx{x}_i^T\mtx{v}_{[k+1:k+P], j}}_{b_{u,i}} +
\underbrace{\mtx{u}_{[k+1:k+Q], i}^T\mtx{y}_j}_{b_{v,j}},
\end{align}
where $\mtx{x}_i \in \mathbb{R}^{P}$ and $\mtx{y}_j \in \mathbb{R}^{Q}$ encodes user/item side information.% with a vector of size $P$ and $Q$ . 

\begin{figure}[t]
\centering
\begin{minipage}{.50\textwidth}
  \centering
\includegraphics[width=0.6\linewidth]{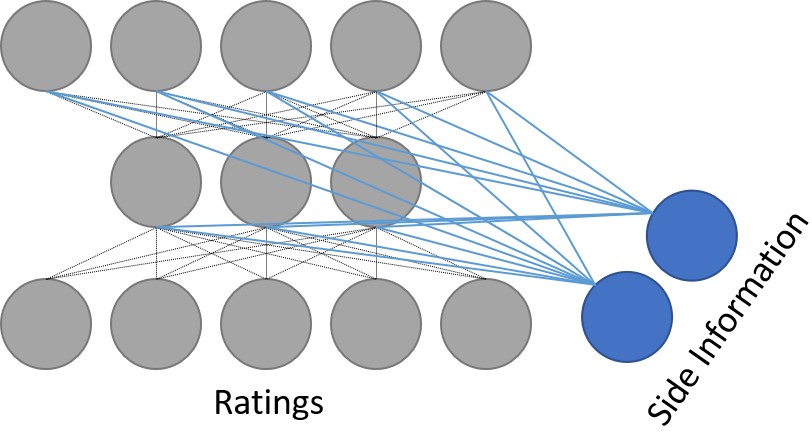}
\caption{Side information is wired to all the neuron.}
\label{fig:side_info}
\end{minipage}
%\quad
%\begin{minipage}{.4\textwidth}
%    \centering
%    \includegraphics[width=0.9\textwidth]{img/converge}
%    \caption{RMSE by epoch for CFN for MovieLens-10M (90\%/10\%). 
%    }
%    \label{fig:converge}
%\end{minipage}
\vskip -0.5em
\end{figure}

Unfortunately, those methods cannot be directly applied to NNs as autoencoders optimize $\mtx{U}$ and $\mtx{V}$ independently. However, it is possible to retrieve similar equations by (i) appending side information to incomplete input vectors, (ii) injecting side information to every layer of the autoencoders as in Fig.~\ref{fig:side_info}. By example, with U-CFN we get the following output:
\begin{align*}
nn(\{ \usparse, \mtx{x}_i\})&= \sigma(\mtx{V}' \;
\{\overbrace{\sigma( \mtx{W}'_1 \{\usparse, \mtx{x}_i\} + \mtx{b_1} )}^{\mtx{u'}_i}, \mtx{x}_i\}  + \mtx{b_2}) \\
%&= \mtx{V}' \; \{ \mtx{u'}_i, \mtx{x}_i\} + \mtx{b_2} \\
&= \sigma(\mtx{V}'_{[1:k]}\mtx{u'}_i +  \underbrace{\mtx{V}'_{[k+1:k+P]}\mtx{x}_i}_{\mtx{b}_{u,i}} + \underbrace{\mtx{b_2}}_{[b_{v,1}\dots b_{v,M}]^T}), \numberthis \label{eq:side_info2} 
\end{align*}
where $\mtx{V}' \in \mathbb{R}^ {(N \times k+P)}$ is a weight matrix, $\mtx{V}'_{[1:k]} \in \mathbb{R}^ {N \times k}, \mtx{V}'_{[k+1:k+P]} \in \mathbb{R}^ {N \times P}$ are respectively the submatrices of $\mtx{V}'$ that contain the columns from $1$ to $k$ and $k+1$ to $k+P$. 
Injecting side information to the last hidden layer as in Eq.~\ref{eq:side_info2} enables to partially retrieve the error function of classic hybrid systems described in Eq.~\ref{eq:side_info1}. %Only the user/item bias is missing from I-CFN/U-CFN. 
Secondly, injecting side information to the other intermediate hidden layers enhance the internal representation. Finally, appending side information to the input vectors supplants the absence of input data when no rating is available. The autoencoder fits CF standards while being trained end-to-end by backpropagation. 

\section{Experiments}
\label{sec:exp}
In this section, we empirically evaluate CFN on two major CF datasets: MovieLens and Douban. We first describe the experimental settings before introducing the benchmark models. Finally, we provide an extensive analysis of the results.

\subsection{Experimental Setting}
Experiments are conducted on MovieLens and Douban datasets as they are among the biggest CF dataset freely available at the time of writing. Besides, MovieLens contains side information on the items and it is a widespread benchmark while Douban provides side information on the users. The MovieLens-1M, MovieLens-10M and Movie\-Lens-20M  respectively provide 1/10/20 million discrete ratings from 6/72/138 thousands users on 4/10/27 thousands movies. Side information for MovieLens-1M is the age, sex and gender of the user and the movie category (action, thriller etc.). Side information for MovieLens-10/20M is a matrix of tags $\mtx{T}$ where $T_{ij}$ is the occurrence of the $j^{th}$ tag for the $i^{th}$ movie and the movie category. No side information is provided for users. The Douban dataset \cite{Hao2011} provides 17 million discrete ratings from 129 thousand users on 58 thousands movies. Side information is the bi-directional user/friend relations for the user. The user/friend relations are treated like the matrix of tags from MovieLens. No side information is provided for items.
Finally, we perform 90\%-10\% train-test sets as reported in \cite{Lee2013,Li2016,Zheng2016}.

\paragraph{Preprocessing}
For each dataset, we first centered the ratings to zero-mean by row (resp. by col) for U-CFN (resp I-CFN): denoting $b_{i}$ the mean of the $i^{th}$ user and $b_{j}$ the mean of the $j^{th}$ item, U-CFN and I-CFN respectively learn from $r_{ij}^{unbiased} = r_{ij} -  b_{i}$ and $r_{ij}^{unbiased} = r_{ij} -  b_{j}$. 
Then all the ratings are linearly rescaled from -1 to 1 to fit the output range of the autoencoder transfer functions.
Theses operations are finally reversed while evaluating the final matrix.

\paragraph{Side Information}
As the dimensionality of the side information is huge, we first perform a dimension reduction of the matrix of tags/friendships. To do so, we use a low rank matrix factorization in our experiments. For a different nature of data as texts or pictures (which are not available in our dataset), CFN can directly learn this embedding as \cite{Wang2014b,Wang2014,Li2015}.
%As the dimensionality of the side information is huge, we first use a dimension reduction.  Many  techniques could be used to perform this step. In theses experiments due to the nature of the available side info this is done using a low rank matrix factorization. For different natures of data such as texts or pictures, one may directly learn this embedding as \cite{???}.
Formally, we use the left part of a matrix factorization of the tag matrix $\mtx{T}$. From  $\mtx{T} = \mtx{P} \mtx{D} \mtx{Q}^T$ with $\mtx{D}$ the diagonal matrix of eigenvalues sorted in descending order, the movie tags are represented by $\mtx{Y} = \mtx{P}_{J\times K'} \mtx{D}_{K'\times K'}^{0.5}$ with $K'$ the number of kept eigenvectors. Binary representation such as the movie category is concatenated to $\mtx{Y}$.

\paragraph{Error Function}
The algorithms are compared based on their respective Root Mean Square Error (RMSE) on test data.
% We reversed the preprocessing operation CFNThe comparisons are performed on the initial data without . 
Denoting $\mtx{R^{test}}$ the matrix of test ratings and $\mtx{\widehat{R}}$ the full matrix returned by the learning algorithm, the RMSE is:
\begin{equation}
L(\mtx{\widehat{R}}, \mtx{R^{test}}) = \sqrt{\frac{1}{|\mathcal{K}(R^{test})|}\sum_{(i,j)\in\mathcal{K}(R^{test})}(r_{ij}^{test}-\hat{r}_{ij})^2},
\end{equation}
where $|\mathcal{K}(R^{test})|$ is the number of ratings in the testing dataset.
Note that in the case of autoencoders $\mtx{\widehat{R}}$ is computed by feeding the network with training data. As such, $\hat{r}_{ij}$ stands for $nn(\mtx{\hat{r}_{i,.}}^{train})_j$ for U-CFN, and  $nn(\mtx{\hat{r}_{.,j}}^{train})_i$ for I-CFN.

\begin{table*}[t]
\centering
\begin{tabular}{lcccc}
\hline
Algorithms  & MovieLens-1M                     & MovieLens-10M             & MovieLens-20M    & Douban\\
\hline
BPMF		&         0.8705  $\pm$ 4.3e-3  &          0.8213 $\pm$ 6.5e-4  &           0.8123 $\pm$ 3.5e-4     & 0.7133 $\pm$ 3.0e-4 \\
ALS-WR      &         0.8433  $\pm$ 1.8e-3  &          0.7830 $\pm$ 1.9e-4  &           0.7746 $\pm$ 2.7e-4     & 0.7010 $\pm$ 3.2e-4 \\
SVDFeature  &         0.8631  $\pm$ 2.5e-3  &          0.7907 $\pm$ 8.4e-4  &           0.7852 $\pm$ 5.4e-4     &              *      \\
LLORMA		&         0.8371  $\pm$ 2.4e-3  &          0.7949 $\pm$ 2.3e-4  &          0.7843 $\pm$ 3.2e-4      & 0.6968 $\pm$ 2.7e-4   \\
I-Autorec   &         \textbf{0.8305 $\pm$ 2.8e-3} &   0.7831 $\pm$ 2.4e-4  &          0.7742 $\pm$ 4.4e-4      & 0.6945 $\pm$ 3.1e-4 \\
\hline
U-CFN       &         0.8574  $\pm$ 2.4e-3  &          0.7954 $\pm$ 7.4e-4  &           0.7856 $\pm$ 1.4e-4     & 0.7049 $\pm$ 2.2e-4 \\
U-CFN++     &         0.8572  $\pm$ 1.6e-3  &                N/A            &                 N/A               & 0.7050 $\pm$ 1.2e-4 \\
I-CFN       &         0.8321  $\pm$ 2.5e-3  &          0.7767 $\pm$ 5.4e-4  &           0.7663 $\pm$ 2.9e-4     & \textbf{0.6911 $\pm$ 3.2e-4} \\
I-CFN++     &         0.8316  $\pm$ 1.9e-3  &  \textbf{0.7754 $\pm$ 6.3e-4} &   \textbf{0.7652 $\pm$ 2.3e-4}       &           N/A      \\
\hline
\end{tabular}
\caption{RMSE on MovieLens-10M (90\%/10\%). The ++ suffix denotes when side information is added to CFN.% N/A is used no side information is available.
}
\label{tab:RMSE}
\end{table*}

\paragraph{Training Settings}
We train a one-hidden layer autoencoders with hyperbolic tangent transfer functions. The layers have $600$ hidden neurons. Weights are randomly initialized  with a uniform law
$\mtx{W_{ij}} \sim {\cal U}\left[-1/\sqrt{n}, 1/\sqrt{n}\right]$.
%$\mtx{W_{ij}} \sim {\cal U}\left[-\frac{1}{\sqrt{n}}, \frac{1}{\sqrt{n}}\right]$.
The latent dimension of the low rank matrix of tags/friendships is set to 50. Hyperparamenters were are fine-tuned by a genetic algorithm and the final learning rate, learning decay and weight decay are respectively set to $0.7$, $0.3$ and $0.5$. $\alpha$, $\beta$ and masking ratio are set to $1$, $0.5$ and $0.25$.  %already used by \cite{Mary2007} in a different context. %(see supplementary material). 
\paragraph{Source code} 
In order to ensure easy reproducibility and reuse the experimental results, we provide the code in an out-of-the-box tutorial in Torch. 
\footnote{
%\url{https://github.com/fstrub95/Autoencoders_cf}
\url{https://github.com/fstrub95/Autoencoders_cf}
}.

\subsection{Benchmark Models}
We benchmark CFN with five matrix completion algorithms:
\begin{itemize}%[topsep=0pt]
\item
{\bf ALS-WR} (Alternating Least Squares with Weighted-$\lambda$-Regularization) \cite{Zhou2008} solves the low-rank MF problem by alternatively fixing $\mtx{U}$ and $\mtx{V}$ and solving the resulting linear regression problem. Experiments are run with the Apache Mahout Software \footnote{\url{http://mahout.apache.org/}} with a rank of 200;
 \item {\bf SVDFeature} \cite{Chen2012} learns a feature-based MF : side information is used to predict the bias term and to reweight the matrix factorization. We use a rank of 64 and tune other hyperparameters by random search;
  \item {\bf BPMF} (Bayesian Probabilistic Matrix Factorization)  infers the matrix decomposition after a statistical model. As a bayesian algorithm, the performances can be improved by the fine tuning of priors over the parameters %but the tuning is prone to overfitting. 
  Here, we use the recommendations of \cite{Salakhutdinov2008} for priors and rank (set to 10).%- ensuring that initialization respects mean and variance of rows and columns - . %Note that by casting the regularization used by ALS-WR into priors - which is not straightforward - it should be possible to obtain a better RMSE.
 \item {\bf LLORMA}  estimates the rating matrix as a weighted sum of low-rank matrices. Experiments are run with the Prea API\footnote{\url{http://prea.gatech.edu/}}. We use a rank of 20, 30 anchor points which entail a global pseudo-rank of 600. Other hyperparameters are picked as recommended in \cite{Lee2013};
 \item {\bf I-Autorec} \cite{Sedhain2015} trains one autoencoder per item, sharing the weights between the different autoencoders. We use 600 hidden neurons with the training hyperparameters recommended by the author.
\end{itemize}
In every scenario, we select the highest possible rank which does not lead to overfitting despite a strong regularization. For instance, increasing the rank of BPMF does not significantly increase the final RMSE, idem for SVDFeature. Similar benchmarks exist in the literature \cite{Lee2013,Li2016,Zheng2016}.

\subsection{General Results} 
\begin{figure*}[t]
\centering
\begin{minipage}{.60\textwidth}
	\centering
	\includegraphics[width=0.92\textwidth]{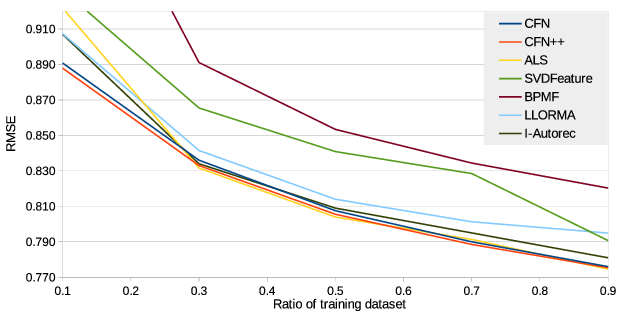}
	\caption{RMSE vs training set ratio for MovieLens-10M. Training hyperparameters are kept constant across dataset. CFN and I-Autorec are very robust to a change in the density while SVDFeature must be refined each time.}
	\label{fig:trainingRatio}
\end{minipage}
\quad
\begin{minipage}{.35\textwidth}
    \centering
    \includegraphics[width=1\textwidth]{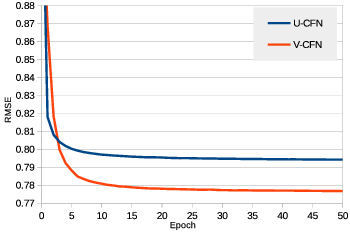}
    \caption{RMSE by epoch for CFN for MovieLens-10M (90\%/10\%). 
    }
    \label{fig:converge}
\end{minipage}
%\vskip -0.5em
\end{figure*}

\paragraph{Comparison to state-of-the-art}
Tab.~\ref{tab:RMSE} summarizes the RMSE on MovieLens and Douban datasets. Confidence intervals correspond to a 95\% range. I-CFNs have excellent performance for every dataset we run. It is competitive compared to the state-of-the-art CF algorithms and outperforms them for MovieLens-10M. To the best of our knowledge, the best result published for MovieLens-10M (without side information) has a final RMSE of $0.7682$~\cite{Li2016}. %From the best of our knowledge, no recent works manage to both reached state of the art results while successfully integrating side information. 
Note that I-CFN outperforms U-CFN as shown in Fig.~\ref{fig:converge}. It suggests that the structure of the items is stronger than the one on the users \textit{i.e.} it is easier to guess tastes based on movies you liked than to find users similar to you. Yet, this behavior could be different on another dataset. Finally, I-CFN outperforms Autorec on big dataset while both methods rely on autoencoder. In addition to the DAE loss, CFN benefits from regularizing missing ratings during the training as described in Eq.~\ref{eq:loss}. Thus, uncommon ratings are more regularized and they turn out to be less overfitted.
\vspace{-0.5em}
\paragraph{Impact of side information}
At first sight at Tab.~\ref{tab:RMSE}, the use of side information has a limited impact on the RMSE. This statement has to be mitigated: as the repartition of known entries in the dataset is not uniform, the estimates are biased towards users and items with a lot of ratings. For theses users and movies, the dataset already contains a lot of information, thus having some extra information will have a marginal effect. Users and items with few ratings should benefit more from some side information but the estimation bias hides them. 
In order to exhibit the utility of side information, we report in Tab.~\ref{tab:Side} the RMSE conditionally to the number of missing values for items. As expected, the fewer number of ratings for an item, the more important is the side information. This is very desirable for a real system: the effective use of side information to the new items is crucial to deal with the flow of new products. A more careful analysis of the RMSE improvement in this setting shows that the improvement is uniformly distributed over the users whatever their number of ratings. This corresponds to the fact that the available side information is only about items. Finally, we train I-CFN on MovieLens-10M (90\%/10\%) with either the movie genre or the matrix of tags. Individually picked, side information increases the global RMSE by 0.10\% while concatenating them increases the final score by 0.14\%. Therefore, I-CFN handles the heterogeneity of side information.

 \begin{table}[t]
\centering
\begin{subtable}{1\linewidth}
\centering

\begin{tabular}{lccc}
Interval \;& I-CFN & I-CFN++ & \%Improv. \\
\hline
0.0-0.2 &  1.0030   &  0.9938  & 0.96 \\
0.2-0.4 &  0.9188   &  0.9084  & 1.15 \\
0.4-0.6 &  0.8748   &  0.8669  & 0.91 \\
0.6-0.8 &  0.8473   &  0.8420  & 0.63 \\
0.8-1.0 &  0.7976   &  0.7964  & 0.15 \\
\hline
Full    &  0.8075   &  0.8055  & 0.25 \\
\hline
\end{tabular}
\caption{MovieLens-10M (50\%/50\%)}
\end{subtable}%
\vskip 1em
\centering
\begin{subtable}{1\linewidth}
\centering
\begin{tabular}{lccc}
Interval \;& I-CFN & I-CFN++ & \%Improv. \\
\hline
0.0-0.2 &  0.9539  &  0.9444  & 1.01 \\
0.2-0.4 &  0.8815  &  0.8730  & 0.96 \\
0.4-0.6 &  0.8487  &  0.8408  & 0.95 \\
0.6-0.8 &  0.8139  &  0.8110  & 0.35 \\
0.8-1.0 &  0.7674  &  0.7669  & 0.06 \\
\hline
Full    &  0.7767   &  0.7756  &  0.14 \\
\hline
\end{tabular}
\caption{MovieLens-10M (90\%/10\%)}
\end{subtable}
\caption{RMSE computed by clusters of items sorted by the number of ratings on MovieLens-10M (90\%/10\%). For instance, the first cluster contains the 20\% of items with the lowest number of ratings.}
\label{tab:Side}
\vskip -1em
\end{table}

%\begin{table}[h]
%\caption{Impact of the denoising loss in the training process. If we focus on the prediction, the %autoencoder provides poor results. If we focus on the reconstruction with no masking noise, the %autoencoder already provides excellent results. By mixing them, the network converges to a better %score.}	    	    	    
%\label{tab:DAEimpact}

%\caption{MovieLens-10M (90\%/10\%)}
%\begin{tabular}{c|lccc|c}
%& & \; $\alpha$  \;& \;$\beta$  \; & \%Mask & RMSE \\
%	\hline
%			&Superv.   &  0.91     & 0     & 0    & 0.8020\\
%MLens-10M	&Unsup.    &  0        & 0.54  & 0.25 & 0.7795\\
%			&Mixed     &  0.91     & 0.54  & 0.25 & 0.7768\\
%	\hline
%			&Superv.   &  1        & 0     & 0.25 & 0.7982\\
%MLens-10M	&Unsup.    &  0        & 0.60  & 0    & 0.7690\\
%			&Mixed     &  1        & 0.60  & 0.25 & 0.7663\\
%	\hline
%\end{tabular}
%\end{table}

\begin{figure}[t]
\centering
\includegraphics[width=0.40\textwidth]{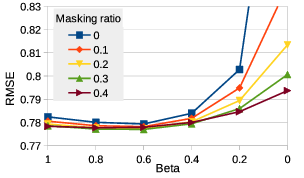}
	\caption{Impact of the denoising loss in the training process. $\alpha = 1$ is kept constant while varying the reconstruction hyperparameter $\beta$ and the masking ratio .}
	\label{fig:DAEimpact}
    \vskip -0.5em
\end{figure}

\vspace{-0.5em}
\paragraph{Impact of the loss}
The DAE loss positively impacts the final RMSE on big dataset as highlighted in Fig.~\ref{fig:DAEimpact} when carefully balanced. Surprisingly, the autoencoder has already good generalization properties while only focusing on the reconstruction criterion (no masking). More importantly, the reconstruction criterion cannot be discarded  ($\beta=0$) to learn an efficient representation. %However, the DAE loss is disadvantageous on small dataset as entries are less redundant. 
\vspace{-0.5em}
\paragraph{Impact of the non-linearity}
We removed the non-linearity from I-CFN to study its relative impact. For fairness, we kept $\alpha$, $\beta$, the masking ratio and the number of hidden neurons constant. We fine-tune the learning rates and weight decay. For MovieLens-10M, we obtain a final RMSE of 0.8151 $\pm$ 1.4e-3 which is far worse than classic non-linear I-CFN. 
\vspace{-0.5em}
\paragraph{Impact of the training ratio}
CFN remains very robust to a variation of data density as shown in Fig.~\ref{fig:trainingRatio}. It is all the more impressive that hyperparameters are first optimized for movieLens-10M (90\%/10\%). Cold-start and warm-start scenario are also far more well-handled by NNs than more classic CF algorithms. 

%Finally, CFN turns out to an efficient and robust algorithm for CF tasks. %These are highly valuable properties in an industrial context.

\begin{table}[h]
\centering 	    
%\caption{U-CFN}
\begin{tabular}{clrrr}
\hline
Model & Dataset & \# Param & Time & Memory  \\
\hline
\multirow{4}{*}{U-CFN} 
 & MLens-1M    &  5M  &  7m17s  &  262MiB  \\
 & MLens-10M   & 15M  & 34m51s  &  543MiB  \\
 & MLens-20M   & 38M  & 59m35s  & 1,044Mib  \\
\hline	
\multirow{4}{*}{I-CFN} 
 & MLens-1M   &   8M &  2m03s  &  250MiB  \\
 & MLens-10M   & 100M & 18m34s  & 1,532MiB  \\
 & MLens-20M   & 194M & 34m45s  & 2,905MiB  \\
\hline	
\end{tabular}
\caption{Training time and memory footprint for a 2-layers CFN on GTX 980 for 20 epochs.%Adding side information would increase by  than 5\% the final time and memory footprint.%
}
\label{tab:Time}
\end{table}

\subsection{CFN Tractability}
One major problem faced by CF algorithms is scalability as CF datasets contain hundred of thousands of users and items.
% An efficient algorithm must be trained in a reasonable amount of time and provide quick feedback during evaluation time.
%
Recent advances in GPU computation managed to reduce the training time of NNs by several orders of magnitude. CFN (and ALS-WR) fully benefits from those advances and is trained within a few minutes as shown in Tab.~\ref{tab:Time}.  On the other side, CF gradient based methods such as SVDFeature \cite{Chen2012} cannot be easily parallelized or used on GPU as they are mainly iterative. Similarly, Autorec \cite{Sedhain2015}  suffers from  synchronization and memory fetching latencies because of the shared weights among autoencoders. 
Furthermore, CFN has the key advantage to provide excellent performance while being able to refine its prediction on the fly for new ratings. Thus, U-CFN/I-CFN does not need to be retrained for new items/users. 

\section{Conclusion} 
In this paper, we highlight the connections between autoencoders and matrix factorization for matrix completion. We pack some modern training techniques - as well than some code - able to defeat state of the art methods while remaining scalable.   
Moreover, we propose a systematic way to integrate side information without the need to combine two separate systems. To some extent, this work extends the construction of embeddings by neural networks to the collaborative filtering setting. A natural follow-up is to work with deeper architectures using batch normalization \cite{Ioffe2015}, adaptive gradient methods (such as ADAM \cite{Kingma2014})
or residual networks \cite{He2015}. Other extensions could use recurrent networks to grasp the sequential aspect of the collection of ratings.

\subsubsection*{Acknowledgements}
The authors would like to acknowledge the stimulating environment provided by SequeL research group, Inria and CRIStAL. This work was supported by French Ministry of Higher Education and Research, by CPER Nord-Pas de Calais / FEDER DATA Advanced data science and technologies 2015-2020, the Projet CHIST-ERA IGLU and by FUI Herm\`{e}s. Experiments presented in this paper were carried out using Grid'5000 testbed, hosted by Inria and supported by CNRS, RENATER and several Universities as well as other organizations. 

{\small
\bibliographystyle{named}
\bibliography{biblio}
}
\end{document}